\documentclass[conference]{IEEEtran}
\IEEEoverridecommandlockouts
\usepackage{cite}
\usepackage{amsmath,amssymb,amsfonts}
\usepackage{graphicx}
\usepackage{textcomp}
\usepackage{xcolor}
\usepackage{url}
\usepackage[capitalise]{cleveref}
\usepackage{tikz}
\usepackage{pgfplots}
\usepackage{algorithm}
\usepackage{algpseudocode}
\usepackage{booktabs}

\newcommand{\nds}{$1032$}
\newcommand{\mname}{\textbf{TSMS}}
\DeclareMathOperator*{\argmin}{arg~min}
\DeclareMathOperator*{\E}{\mathbb{E}}
\newcommand{\T}[2]{T^{(#1:#2)}}
\newcommand{\X}[1]{X^{(#1)}}

\newtheorem{definition}{Definition}

\def\BibTeX{{\rm B\kern-.05em{\sc i\kern-.025em b}\kern-.08em
    T\kern-.1667em\lower.7ex\hbox{E}\kern-.125emX}}
\begin{document}

\title{Explainable Adaptive Tree-based Model Selection for Time-Series Forecasting}

\author{\IEEEauthorblockN{Matthias Jakobs}
\IEEEauthorblockA{\textit{Lamarr Institute for Machine Learning } \\
\textit{and Artificial Intelligence} \\
TU Dortmund University \\
Dortmund, Germany \\
matthias.jakobs@tu-dortmund.de}
\and
\IEEEauthorblockN{Amal Saadallah}
\IEEEauthorblockA{\textit{Lamarr Institute for Machine Learning } \\
\textit{and Artificial Intelligence} \\
TU Dortmund University \\
Dortmund, Germany \\
amal.saadallah@cs.tu-dortmund.de}

}

\maketitle

\begin{abstract}
Tree-based models have been successfully applied to a wide variety of tasks, including time series forecasting.
They are increasingly in demand and widely accepted because of their comparatively high level of interpretability. However, many of them suffer from the overfitting problem, which limits their application in real-world decision-making. This problem becomes even more severe in online-forecasting settings where time series observations are incrementally acquired, and the
distributions from which they are drawn may keep changing over time. In this context, we propose a novel method for the online selection of tree-based models using the TreeSHAP explainability method in the task of time series forecasting. We start with an arbitrary set of different tree-based models. Then, we outline a performance-based ranking with a coherent design to make TreeSHAP able to specialize the tree-based forecasters across different regions in  the input time series. In this framework, adequate model selection is performed online, adaptively following drift detection in the time series. In addition, explainability is supported on three levels, namely online input importance, model selection, and model output explanation. An extensive empirical study on various real-world datasets demonstrates that our method  achieves excellent or on-par results in comparison to the state-of-the-art approaches as well as several baselines.
\end{abstract}

\begin{IEEEkeywords}
Online Model Selection, Tree-based Models, Time Series Forecasting, TreeSHAP, Explainability
\end{IEEEkeywords}

\section{Introduction}
Time series forecasting has always been recognized as a main component of informed decision-making across various real-world applications, including but not limited to smart manufacturing control, predictive maintenance, energy management, transport planning in smart cities, and financial investments \cite{godahewa2021monash,hyndman/etal/2008a,Saadallah/Moreira/2018a}. However, due to the complex and dynamic nature of time series data, forecasting is often considered to be one of the most challenging tasks in time series analysis. Time series data can contain sources of non-stationary variations and are therefore susceptible to the concept drift phenomenon \cite{gama2014survey}, which further adds to the complexity of forecasting.
There have been numerous Machine Learning (ML) models suggested for addressing this task. One option involves handling the data as sequential observations either in an online or streaming fashion \cite{hyndman/etal/2015a}. Another approach is to utilize time series embeddings that convert a group of target observations into a feature space of $k$ dimensions corresponding to the observation's past $k$-lagged values \cite{saadallah/etal/2019,cerqueira/etal/2017a}. 
%No single best model

Although various ML forecasting methods have been proposed, it is widely recognized that none of them can universally apply to all applications. This limitation is a direct consequence of the No Free Lunch theorem established by Wolpert \cite{wolpert/1996}, which states that no learning algorithm can perform optimally for all learning tasks. Furthermore, even within the same application, models exhibit varying performance over time \cite{saadallah/etal/2019,saadallah/etal/2020,cerqueira/etal/2017a,priebe2019}, which can be attributed to the aforementioned difficulties in time series modeling \cite{saadallah/jacobs/21,saadallah/etal/2022a,saadallah/etal/2019}. It can also be explained by the fact that different forecasting models have expected areas of expertise or competence, referred to as Regions of Competence (RoCs), distributed across different regions of the input time series, which can lead to varying relative performance in these regions \cite{saadallah/jacobs/21,priebe2019,saadallah/etal/2022a}.
The nature of the forecasting application can provide a guideline for selecting the adequate family of ML models to be applied. With the increasing number of safety-critical scenarios, model explainability is becoming more and more required \cite{molnarInterpretableMachineLearning2020}. In this case, interpretable models per construction, such as tree-based models, are favored \cite{taieb2014gradient,ilic2021explainable}. Still, adequate model selection has to cope with the time-evolving nature of time series data that may be subject to the concept drift phenomenon. However, most of the existing tree-based models, including Decision Trees and
their ensembles, such as Random Forests and Gradient-boosted Trees, are restricted to operate in a static manner in the ML literature, i.e., they do not take into account changes in the time series \cite{breiman2001random,friedman2001greedy,taieb2014gradient,galicia2019multi}.
One possible way to mitigate this issue is either to retrain them periodically in a blind manner, i.e., without any knowledge or assumption about the presence of concept drift, or to apply them together with a drift detection mechanism that triggers their update in an informed manner only when necessary \cite{saadallah/etal/2019,saadallah/jacobs/21,saadallah/etal/2022a}.

More recently, the concept of \textit{Regions of Competence} (RoCs) has been used for model selection \cite{saadallah/jacobs/21,cerqueira/etal/2017a,priebe2019,cerqueira/etal/2018a,saadallah/etal/2022a} in the forecasting task. Different ways to determine these RoCs have been proposed \cite{cerqueira/etal/2017a,priebe2019,saadallah/jacobs/21}, including model-type-independent approaches that use meta-learning by either training meta-models to predict the performance of candidate models on the most recent time series pattern of $k$-lagged values \cite{priebe2019} or at a particular test time \cite{cerqueira/etal/2018a,cerqueira/etal/2017a}. The selection approaches in these works are performed online in a blind manner at each time step, i.e., without taking into account the occurrence of significant changes in either the time series data or the performance of the candidate models. 
In \cite{saadallah/jacobs/21}, a model-specific approach, in particular for Deep Neural Networks (DNNs), was developed to compute RoCs. 
Although the RoCs are updated in an informed manner following a concept drift in the time series, the proposed method is specific to a particular class of DNNs, namely Convolutional Neural Networks (CNNs), and cannot be generalized to any family of forecasting models as it is based on Grad-CAM \cite{saadallah/jacobs/21}, a gradient-based heatmapping method for highlighting important input regions in CNNs.
The advantage of the developed method is that it provides some visual explanations for the reason for selecting a particular CNN at a particular test time. 
Even though the model selection process is made explainable via heatmaps generated by Grad-CAM, the candidate CNNs themselves are considered black-box models that suffer from lack of interpretability and transparency, thus limiting their applicability in many real-world scenarios and their acceptance by end-users \cite{liang2021explaining,saadallah2020active}.

In this paper, we propose an explainable online adaptive tree-based model selection strategy for time series forecasting. The selection among candidate tree-based models is based on the concept of RoCs, which are determined using TreeSHAP, which is a tree-based method for explaining individual predictions by computing the contribution of each feature to the prediction method in the form of Shapley values \cite{lundbergLocalExplanationsGlobal2020}. We devise a coherent design to make TreeSHAP able to specialize the tree-based forecasters across different regions in the input time series using a performance-based ranking over a time-sliding window validation set. At each time step, the distances between the recently observed window of time series observations (i.e., lagged values used to compute the forecast) and the pre-computed RoCs are determined. The model corresponding to the RoC with the lowest distance is selected to perform the forecasting. At test time, to account for the potential emergence of new concepts in the data, the pre-calculated RoCs are dynamically updated when a concept drift is identified in the time series. This is achieved by shifting the validation set, thereby enabling the computation of RoCs that reflect the changing nature of the data. In addition, our method supports explainability on three levels, namely, input time series importance, model selection, and model output explanation. 

We carried out an extensive empirical evaluation of our framework by applying it to \nds\ real-world time series datasets from diverse domains. The results obtained indicate that our method outperforms state-of-the-art techniques for online model selection and several baselines for time series forecasting. It is worth mentioning that our experiments are entirely reproducible, and we have made both the code and how to access the datasets publicly available\footnote{\scriptsize\url{https://github.com/MatthiasJakobs/tsms}}.
The main contributions of this paper are thus summarized as follows:
\begin{itemize}
	\item We present a novel method for online tree-based-model selection for time series forecasting by computing candidate models' RoCs using an adaption of the TreeSHAP method.
	\item We update the  RoCs in an informed manner following  concept drift detection in the time series data.
	\item We exploit the RoCs to support explainability on three levels, namely, input time series selection, model selection, and model output explanation.
	\item We provide a comparative empirical study with state-of-the-art methods and discuss their implications in terms of predictive performance and scalability.
\end{itemize}

\section{Related Work}
\label{sec:lit}
Online single model selection has been revealed to be challenging in the context of time series forecasting due to the dynamic nature of the data \cite{saadallah/etal/2019} and the resulting time-evolving models' performance/competence \cite{saadallah/jacobs/21}. Several techniques have been proposed to choose the appropriate model from a set of candidate models for a specific forecasting task. These techniques can be classified into three primary families. The first family involves approximating a posterior distribution over each candidate's expected error using parametric or non-parametric estimation methods, such as Gaussian approximation \cite{birge2001gaussian} or Bayesian estimation \cite{argiento2010bayesian}. However, these methods are not practical in forecasting contexts as they require approximating continuous composite densities for the error based on target and estimated time series values.
The second family of selection methods involves estimating the unseen error of a given model through empirical evaluations, such as using an independent validation or calibration dataset. Models with the lowest estimated error are subsequently chosen \cite{rivals1999cross}. However, the empirical error may in many cases not serve as an accurate estimate of the true error \cite{shalev2014understanding}.
The third family of methods is based on the meta-learning paradigm, where the selection of the appropriate method is decided by another machine learning model that learns from previous selection realizations characterized by a set of devised meta-features~\cite{wolpert1992stacked}. Meta-learning has been established as a useful tool for model selection in time series forecasting \cite{cerqueira/etal/2017a,saadallah/etal/2019,saadallah/jacobs/21}. 

More recently, the concept of \textit{Regions of Competence} (RoCs) has been used both for the selection of a single model \cite{saadallah/jacobs/21} and for the selection of ensemble members \cite{cerqueira/etal/2018a,cerqueira/etal/2017a,priebe2019} in the forecasting task. 
In \cite{priebe2019}, at test time, the most similar pattern to the current input (i.e., in our case, time series input sequence) is determined, and the model with the smallest error is selected for prediction.  
In \cite{cerqueira/etal/2018a,cerqueira/etal/2017a},
 meta-learning is used to build models capable of modeling the competence of each of the ensemble members across the input space.
The authors frame their ensemble learning as a ranking task, in which ensemble members are ranked sequentially by their decreasing weight (i.e., the one predicted to perform better is ranked first). Correlation among the output of the base learners is used to quantify their redundancy. A given learner is penalized for its correlation to each learner already ranked. If it is fully correlated with other learners already ranked, its weight becomes zero. Opposingly, if it is completely uncorrelated with its ranked peers, it gets ranked with its original weight. 
In the above methods, the meta-models responsible for computing the RoCs are kept static over time. Only distances or error comparisons are performed online in a blind-manner at each time step, i.e., without taking into account the occurrence of significant changes in either the time series data or the performance of the candidate models. 
%As mentioned in the introduction, the recent method proposed in \cite{saadallah/jacobs/21} for computing RoCs, even though adaptive, it is CNNs-specific. The provided model selection explanations depends on some computed gradient-based saliency map and therefore are also CNNs specific. 

In \cite{saadallah/jacobs/21}, RoCs are determined using performance gradient-based saliency maps, which establish a mapping between the input time series and the performance of the DNNs. A drift detection mechanism in the data is deployed to update the RoCs. 
These maps are inspired by class activation maps (Grad-CAM) \cite{selvaraju2017grad}, which have been widely used as a visual explainability tool in computer vision tasks. More recently, Grad-CAM has been applied in the context of time series classification to explain which time series features in which time intervals are responsible for a given predicted class \cite{assaf2019explainable}.
The authors in \cite{saadallah/jacobs/21} exploit the gradient maps to provide explanations for the reason behind selecting a given CNN at a particular time instant or interval. However, the computation of RoCs and the resulting explanations are specific to CNNs and can not be generalized to any other model. In addition, despite making the model selection process explainable, the candidate CNNs are still considered black-box models that lack interpretability and transparency. This limitation restricts their usability in many real-world situations and their acceptance among end-users \cite{liang2021explaining,saadallah2020active}.

Therefore, in this work, we focus on interpretable models per construction, namely tree-based models, and we leverage the RoCs concept to adapt them to the online forecasting application through an adaptive model selection procedure.

\section{Methodology}

% Intro
Our proposed method builds upon OS-PGSM, presented in \cite{saadallah/jacobs/21}.
First, we will define the used notation.
Second, we describe Shapley values with a focus on TreeSHAP \cite{lundbergLocalExplanationsGlobal2020}.
Third, we will reformulate the earlier method and highlight the changes we propose to allow for the efficient selection of tree-based models.

% Basic notation
\subsection{Preliminaries}
In this work, we will focus on univariate time series forecasting.
Let $X$ be a series of values, with $X^{(t)}$ denoting the value at time step $t$.
We use shift operator notation:
$$
    T^k\X{t} = \X{t+k}
$$
Note that we default to shifting to the right, forwards in time, instead of backward in time.
Additionally, we define the shift operator on intervals as follows:
$$
    \T{u}{v}\X{t} = (T^u \X{t}, T^{u+1} \X{t}, \dots, T^{v}\X{t})
$$
For ease of notation, we further define
$$
\T{}{v} := \T{1}{v}, ~~~ \T{-v}{} := \T{-v}{0}
$$
The goal of forecasting at time $t$ is to predict the next $H$ values $\T{}{H}\X{t}$ from known $L$ lagged values $\T{-L}{}\X{t}$.

\subsection{Shapley values and TreeSHAP}
Shapley values, originally proposed in Game Theory \cite{shapleyValueNpersonGames1953}, have become a popular method for explaining ML models in recent years \cite{rozemberczkiShapleyValueMachine2022}.
They attribute the outcome of a game, given by a value function $v$, onto each player $i$ who participates in this game.
In ML applications, the players are often chosen as the features $\{ i \in N \}$ of a specific data point $x$ with an optional label $y$.

\begin{definition}
    The Shapley value $\phi_i$ with a defined value function $v$ is given by

    \begin{align*}
        \phi_i(v) = \sum_{S \subseteq N \setminus \{ i \}} \binom{|N|-1}{|S|}^{-1}  \frac{v(x, y, S \cup \{ i \}) - v(x, y, S)}{|N|} 
    \end{align*}

\end{definition}

% Axioms
One reason often stated for the use of Shapley values in explaining machine learning models is due to desired axioms for which Shapley values are the only solution\cite{rozemberczkiShapleyValueMachine2022}.
We want to highlight one axiom that we will utilize later, namely \textit{Linearity}, which states that the Shapley value of a linear combination of value functions is equal to the linear combination of the individual Shapley values:
$$
    \phi_i(v_a) + \phi_i(v_b) = \phi_i(v_a + v_b) ~ \forall i \in N
$$

TreeSHAP \cite{lundbergLocalExplanationsGlobal2020} is a method to estimate Shapley values in polynomial time if the models are tree-based.
The method can either compute Shapley values directly via the tree structure (\textit{tree-dependent} approach) or via a held-out set of data points (\textit{interventional} approach).
We use the second approach since the official reference implementation\footnote{\url{https://github.com/slundberg/shap}} allows us to explain not only the prediction but also the squared loss of tree-based models, whereas the \textit{tree-dependent} approach is limited to explaining the prediction. 

TreeSHAP defines, depending on the problem to be explained, two different value functions.
Let $g: \mathbb{R}^N \rightarrow \mathbb{R}$ be a trained regression decision tree, which takes $N$ features as input.
If the decision is to be explained, the value function is defined by
\begin{equation}
    v_{g,pred}(x, S) = \E_{X \sim \mathcal{X}}[g(X ~ | ~ X_S = x_S)]  
\end{equation}
where $\mathcal{X}$ is a set of reference data, often referred to as background data.
We set the background data to be equal to the training data for the trees in our experiments.
The other value function is used if we want to explain the squared loss of a tree's prediction and is defined by
\begin{equation}
    \label{eq:v_loss}
    v_{g,loss}(x, y, S) = \E_{X \sim \mathcal{X}}[(g(X ~ | ~ X_s = x_S) - y)^2]
\end{equation}

TreeSHAP implements an efficient, dynamic programming-based algorithm to compute the Shapley values for one datapoint in polynomial time by going down the tree just once.
By leveraging the Linearity axiom, the computed Shapley values of individual trees in an ensemble can be combined to give the Shapley values of the ensemble \cite{lundbergLocalExplanationsGlobal2020}.

\subsection{TreeSHAP Model Selection (TSMS)}

Next, we present our approach, TreeSHAP Model Selection \textbf{(TSMS)}.
Assume a model pool $\mathcal{P} = \{ f_1, \dots, f_M \}$ of $M$ forecasters from which we want to choose the forecaster $\hat{f}$ that minimizes a loss $\mathcal{L}$ for a window $\T{-L}{}\X{t}$ and its corresponding horizon $\T{}{H}\X{t}$ i.e., 

$$
    \hat{f} = \argmin_{i \in \{ 1, \dots, M \} } \mathcal{L}\left(f_i(\T{-L}{}\X{t}), \T{}{H}\X{t}\right)
$$

Each time series $X$ is partitioned into three subseries $X_{train}$, $X_{val}$ and $X_{test}$.  
The forecasters are first trained on windowed segments of $X_{train}$.
Afterwards, let $\mathcal{R} := \{ \mathcal{R}_1, \dots, \mathcal{R}_M \}$ define each forecasters Region of Competence (RoC) over $X_{val}$.

\textbf{RoC Creation}: A RoC is filled with subseries of data on which the forecaster performed best in comparison to all other forecasters.
%To build each RoC over $X_{val}$, we utilize \cref{alg:rebuild_roc}.
First, we partition $X_{val}$ into $n_\omega$ chunks of size $\omega$:
\begin{align*}
    X_{val, \omega} = \{ ~ &\T{0}{\omega-1}\X{1}, \T{0}{\omega-1}\X{1+\omega}, \dots, \\
    &\T{0}{\omega-1} \X{1+(n_{\omega}-1)\omega} ~ \}
\end{align*}

with corresponding forecasting labels
$$
    Y_{val, \omega} = \{ ~ \T{}{H}\X{\omega}, \T{}{H}\X{2\omega}, \dots, \T{}{H} \X{n_{\omega}\omega} ~ \}
$$
For each pair $(X_{val, \omega}^{k}, Y_{val, \omega}^{k})$ with $k \in \{ 1, \dots, n_\omega \}$, we find the best performing forecaster $\hat{f}_k$ in terms of squared loss $\mathcal{L}$:
%For each pair $(X_{val, \omega}^{(k)}, Y_{val, \omega}^{(k)})$ with $k \in \{ 1, \dots, n_\omega \}$, we find the best performing forecaster $\hat{f}_k$ in terms of squared loss $\mathcal{L}$:

$$
\hat{f}_k = \argmin_{i \in \{ 1, \dots, M \}} \mathcal{L}(f_i(X_{val, \omega}^{k}), Y_{val, \omega}^{k})
$$
% $$
% \hat{f}_k = \argmin_{i \in \{ 1, \dots, M \}} \mathcal{L}(f_i(X_{val, \omega}^{(k)}), Y_{val, \omega}^{(k)})
% $$

Next, we split each $X_{val, \omega}^{k}$ further into windows of size $L$ with a step size of one:
%Next, we split each $X_{val, \omega}^{(k)}$ further into windows of size $L$ with a step size of one:
\begin{align*}
    x_{val,\omega,k} = \{ ~ &\T{0}{L-1} X_{val, \omega}^{k, (1)}, \T{0}{L-1} X_{val, \omega}^{k, (2)}, \dots, \\ 
    &\T{0}{L} X_{val, \omega}^{k, (\omega - L - H +1)}  ~ \} \\
\end{align*}
\begin{align*}
    y_{val,\omega,k} = \{ ~ &\T{}{H} X_{val, \omega}^{k, (L)}, \T{}{H} X_{val, \omega}^{k, (L+1)}, \dots, \\
    &\T{}{H} X_{val, \omega}^{k, (N-H)}  ~ \}
\end{align*}
% \begin{align*}
%     x_{val,\omega,k} &= \{ ~ \T{}{L} X_{val, \omega}^{(k)}, \T{1}{L+1} X_{val, \omega}^{(k)}, \dots, \T{\omega-L-H}{\omega-H} X_{val, \omega}^{(k)}  ~ \} \\
%     y_{val,\omega,k} &= \{ ~ \T{L}{L+H} X_{val, \omega}^{(k)}, \T{L+1}{L+H+1} X_{val, \omega}^{(k)}, \dots, \T{\omega-H}{\omega} X_{val, \omega}^{(k)}  ~ \}
% \end{align*}

Again, we consider each pair $(x_{val,\omega,k}^{p}, y_{val,\omega,k}^{p})$ and compute Shapley values, decomposing the loss $\mathcal{L}(\hat{f}_k(x_{val,\omega,k}^{p}), y_{val,\omega,k}^{p})$ onto each input feature, resulting in explanation $\phi^{p}_{k} \in \mathbb{R}^L$. 
% Again, we consider each pair $(x_{val,\omega,k}^{(p)}, y_{val,\omega,k}^{(p)})$ and compute Shapley values, decomposing the loss $\mathcal{L}(\hat{f}_k(x_{val,\omega,k}^{(p)}), y_{val,\omega,k}^{(p)})$ onto each input feature, resulting in explanation $\phi^{(p)}_{k} \in \mathbb{R}^L$. 
Our aim is to add refined versions of each $x_{val,\omega,k}^{p}$ to the RoC of $\hat{f}_k$ to retain only the most salient parts of the input that reduced the achieved loss the most. 
% Our aim is to add refined versions of each $x_{val,\omega,k}^{(p)}$ to the RoC of $\hat{f}_k$ to retain only the most salient parts of the input that reduced the achieved loss the most. 
Thus, to get a measure of saliency, we need to invert the Shapley values so that large positive values now indicate a large reduction of loss and vice versa.
To refine, we threshold $-\phi^{p}_{k}$ by a positive constant $\tau$, setting every explanation below the threshold to zero.
% To refine, we threshold $-\phi^{(p)}_{k}$ by a positive constant $\tau$, setting every explanation below the threshold to zero.
We then use the indices of consecutive, non-zero subsequences resulting from the thresholding to extract subsequences of $x_{val,\omega,k}^{p}$ if they are longer than $2$.
% We then use the indices of consecutive, non-zero subsequence resulting from the thresholding to extract subsequences of $x_{val,\omega,k}^{(p)}$, if they are longer than $2$.
For example, consider the thresholded explanation $-\phi = (0, 0.5, 0.3, 0, 0.1, 0.2, 1.3)$ for some sequence $x$.
Then, we would extract $x^{(5:7)}$ and add it to the Region of Competence.
We decided to discard subseries of size $2$ and below since they do not reveal clear patterns and thus add a lot of noise to the RoCs.

\textbf{Model selection}: During testing, for each new window $\T{-L}{}\X{t}_{test}$, the distance to all RoC members for all forecasters is evaluated, using Dynamic Time Warping (DTW).
We chose DTW because it allows us to have RoC members of different lengths since DTW does not constrain its inputs to have equal lengths.
The forecaster $\hat{f}$, whose RoC $\mathcal{R}$ contains the datapoint with the smallest DTW distance to $\T{-L}{}\X{t}_{test}$ is chosen to forecast, i.e.,

$$
\hat{f} = \argmin_{i \in \{1, \dots, M \} } ~ \min(\{ DTW(\T{-L}{}\X{t}_{test}, r) ~ | ~ r \in \mathcal{R}_i \})
$$

% Concept drift
Over time, the distribution of data might change, which is a phenomenon known and studied as concept drifts \cite{gama2014survey}.
We want our method to adapt to those changes by enriching the Regions of Competence with novel concepts when we detect such drifts.
To account for drifts, we utilize the same concept drift detection method using Hoeffding bounds presented in \cite{saadallah/jacobs/21}, which monitors the deviation of the mean of the time series \cite{saadallah/etal/2019}.
Let $(t_{start}, t_{end})$ be the first and last timestep of the validation data range inside the entire time series $X$.
Let $\mu_0 = \E[\T{t_{start}}{t_{end}}X]$ be the initial expected value over the data.
Then, for each new timestep $j, j > t_{end}$ to forecast, we compute 
$
\Delta_{j} = | \mu_j - \mu_0 |
$
with $\mu_j = \E[\T{t_{end}+1}{j}X]$.
The Hoeffding bound states that after $W$ observations of a random variable with range $r$ the true mean has diverged from $0$ with probability $\sigma$ if
$$
    \Delta_j > \sqrt{\frac{r^2ln(2/\sigma)}{2W}}
$$
where $r$ encodes the range of the data, which we estimate empirically, and $W = j - t_{end}+1$.
We set the user-defined parameter $\sigma=0.99$ to be confident in concept drifts actually occurring.
Also, after each concept drift, we set $\mu_0 = \mu_j$ and reset $t_{start}$ and $t_{end}$ with the range of $\mu_j$.
Most importantly, we create a new RoC for each forecaster from that data.
Afterward, we enrich the old RoCs with the newly created RoCs to retain both old information as well as adapt to changes in the data.

\section{Experiments}
In this section, we aim to answer the following research questions:
\begin{itemize}
    \item \textbf{Q1:} How does \mname\ perform against state-of-the-art model selection methods for tree-based models and other relevant baselines?
    \item \textbf{Q2:} What is the impact of the concept drift adaptation in terms of performance and computational resources, compared to both static pre-computation and blind periodic recreation of the RoCs?
    \item \textbf{Q3:} How can \mname\ in conjunction with tree-based models serve explainability for time-series forecasting?
\end{itemize}

 \begin{table}
    \centering
    \begin{tabular}{lll}
    \toprule
     Name & Nr. of time series & Min. length\\ 
    \midrule
    M4 & $548$ & $251$\\
    tourism & $100$ & $252$\\
    Australian Electricity Demands & $5$ & $500$\\
    Dominick & $100$ & $251$\\
    Bitcoin & $16$ & $500$\\
    Pedestrian counts & $63$ & $500$\\
    KDD Cup & $100$ & $500$\\
    Weather & $100$ & $500$\\
    \midrule
    Total & $1032$ & \\
    \bottomrule
\end{tabular}
    \caption{Breakdown of the used univariate time-series from the Monash Forecasting Repository \cite{godahewa2021monash}.}
    \label{tab:datasets}
\end{table}

\subsection{Experimental Setup}
We utilize \nds\ univariate datasets from various application domains, including financial, weather, and synthetic data.
These datasets are provided by the Monash Forecasting Repository \cite{godahewa2021monash}.
We process each time-series $X$ by using the first $50\%$ for training, the following $25\%$ for validation, and the remaining $25\%$ for testing.
We normalize the entire time series using mean and standard deviation estimated over the training portion of the time series.
Due to this way of splitting the time-series, we discard series that are shorter than $250$ to allow enough training and validation data.
Additionally, we took a random subsample of length $500$ for series that are longer to limit computation time.
%All splits are normalized using the mean and standard deviation over the training data split.

\begin{table}
    \centering
    
\begin{tabular}{ll}
\toprule
Model family & Hyperparameters \\ 
\midrule

Decision Tree & $d_{max} \in \{4, 8, 16 \}$
\\ 
% Random Forest & \begin{tabular}[x]{@{}l@{}} $d_{max} \in \{\infty, 2, 4 \}$ \\ $ n_{trees} \in \{ 16, 32, 64, 128 \} $ \end{tabular}
Random Forest & $d_{max} \in \{2, 4, 6 \}$ \\ 
 & $ n_{trees} \in \{ 16, 32, 64 \} $ \\ 
Gradient Boosting Trees ~ & $d_{max} \in \{2, 4, 6 \}$ \\ 
& $ n_{trees} \in \{ 16, 32, 64 \} $ \\
\bottomrule

\end{tabular}

    \caption{List of all models in the model selection pool. Through combination of the listed hyperparameters, there are 21 models in the final model pool. }
    \label{tab:single_models}
\end{table}

% Single models 
To create the model pool $\mathcal{P}$, we train a total of 21 tree-based models.
%per time series on the training split.
We set $L=15$ and $H=1$ in our experiments to allow for a larger context in terms of lags.
The single models consist of various parametrizations of Decision Trees, Random Forests, and Gradient Boosting Trees from \texttt{scikit-learn}.
We vary the maximum tree depth $d_{max}$ and (if available) the number of ensemble estimators $n_{trees}$ (details shown in \cref{tab:single_models}).
All experiments have been performed on consumer hardware, namely on a 2022 MacBook Pro.
%A breakdown of the single models and used parameters are shown in \cref{tab:list_of_single_models}.

%\input{tables/single_models}

\subsection{TSMS Setup and Baselines}
To investigate the impact of drift detection on model performance and runtime, we propose a total of three variants of our method: 
\begin{itemize}
    \item \mname: Our described method, using drift detection to adapt the RoCs to changes in the time-series
    \item \mname-St: Static variant where drift-detection is disabled. RoCs are created at the beginning of the forecasting process and remain unchanged over time
    \item \mname-Per: A periodic, blind update of the RoCs. We chose to update the RoCs automatically $10$ times in total over the length of $X_{test}$ at fixed, equally spaced intervals
\end{itemize}

Additionally, there are two hyperparameters in our method that we tuned: $\omega$, the size of larger, equal-size windows during the RoC creation process, and $\tau$, the threshold used when refining the data points before addition to an RoC.
We conducted a hyperparameter search and found $\omega=25$ and $\tau=0.01$ to perform very well.

Next, we shortly describe the baseline methods we compare ourselves against: 
\begin{itemize}
    \item Exponential Smoothing \textbf{ETS} \cite{jainStudyTimeSeries2017} and \textbf{ARIMA} \cite{jainStudyTimeSeries2017} which are included as simple, yet important baselines.
    \item \textbf{KNN-RoC} \cite{priebe2019} computes static RoCs using a validation set as input and the rank of the individual candidates in the pool $\mathcal{P}$  on each interval as labels for a \textit{KNN} classifier, using DTW distance and for single model selection $K=1$, is used for comparison. The \textit{KNN} predicts which candidate should be selected at test time
    \item \textbf{CNN} and \textbf{CNN-LSTM} \cite{romeuTimeSeriesForecastingIndoor2013} are Convolutional Neural Network baselines with either a fully-connected or LSTM-based forecaster after the features are extracted
    \item \textbf{DETS} \cite{cerqueira/etal/2018a} is an extension of SWE \cite{Saadallah/Moreira/2018a}, which is an ensemble method that weights the ensemble members based on past performance. \textbf{DETS} extends this by selecting a subset only and using a smoothing function on the average over recent errors for weighting. We compare against selecting the model with the largest weight
    \item \textbf{ADE}~\cite{cerqueira/etal/2017a,cerqueira/etal/2018a} was recently developed for an online dynamic ensemble of forecasters construction. A meta-learning strategy is used that specializes base models across the time series to determine their RoC (see Section \ref{sec:lit}). However, instead of selecting many models, we select the best-performing model using the same principle. 
\end{itemize}

Because of our use of tree-based model, we want to highlight that we cannot compare ourselves to OS-PGSM \cite{saadallah/jacobs/21} because the authors utilize Grad-CAM \cite{selvaraju2017grad}, which only works for Convolutional Neural Networks.

\subsection{Performance Comparison Results}

\begin{table}
    \centering
    % \begin{tabular}{ll}
% \toprule
%                Method & Avg. Rank ($\pm$ variance) \\
% \midrule
%               ETS~~~~ &           $10.56 \pm 2.93$ \\
%             ARIMA~~~~ &            $9.04 \pm 6.14$ \\
%           KNN-RoC~~~~ &            $7.21 \pm 3.62$ \\
%               CNN~~~~ &            $5.53 \pm 9.46$ \\
%              DETS~~~~ &            $5.42 \pm 6.04$ \\
%               ADE~~~~ &            $5.35 \pm 7.00$ \\
%          CNN-LSTM~~~~ &            $4.89 \pm 9.03$ \\
%       Best-Single~~~~ &            $4.66 \pm 7.07$ \\
% \textbf{TSMS-Per}~~~~ &            $4.64 \pm 6.07$ \\
%  \textbf{TSMS-St}~~~~ &            $4.43 \pm 5.42$ \\
%     \textbf{TSMS}~~~~ &            $4.27 \pm 4.44$ \\
% \bottomrule
% \end{tabular}

\begin{tabular}{lrrll}
\toprule
Method & Avg. Rank & Std Deviation & Wins & Losses \\
\midrule
ETS & 7.14 & 4.53 & 789 (740) & 212 (209) \\
ADE & 7.03 & 3.01 & 687 (601) & 314 (218) \\
ARIMA & 6.46 & 5.04 & 737 (679) & 264 (200) \\
KNN-RoC & 6.36 & 2.71 & 633 (578) & 368 (221) \\
DETS & 6.31 & 2.61 & 641 (596) & 360 (295) \\
CNN & 5.49 & 3.35 & 682 (638) & 319 (314) \\
CNN-LSTM & 4.85 & 2.68 & 602 (518) & 399 (378) \\
TSMS-St & 4.64 & 2.88 & 361 (279) & 640 (427) \\
Best-Single & 4.36 & 2.71 & 582 (502) & 419 (414) \\
TSMS-Per & 4.19 & 2.60 & 334 (240) & 667 (482) \\
TSMS & 4.13 & 2.49 & - (-) & - (-) \\
\bottomrule
\end{tabular}

    \caption{Comparison (in terms of average rank achieved over \nds\ datasets) between our method and the baselines. Best-Single is the model from $\mathcal{P}$ that performed best over all datasets.}
    \label{tab:rankinig_table}
\end{table}

\cref{tab:rankinig_table} shows a comparison between our method, including its different variations, to the previously mentioned baselines over \nds\ datasets.
We computed the average rank achieved after measuring the loss using the RMSE score.
As can be seen, the drift-aware variant of \mname\ achieves the smallest average rank and outperforms other online-selection baselines such as DETS, ADE, and KNN-RoC.
For comparison, we also include the best-performing model from $\mathcal{P}$, denoted as Best-Single, and find that it also outperforms the model-selection baselines, indicating that they are not able to predict the performance of this single model on the test data very well.
\mname\ is also able to outperform CNN-LSTM and CNN while being more interpretable at the same time due to our limitations of utilizing depth-restricted tree-based models. 
We also ran a signed Wilcoxon rank test \cite{benavoli2016should} on the wins and losses and indicate the significant wins/losses in paranthesis (significance level 0.05).
The results clearly answer research question \textbf{Q1}.

% ABLATION AND RUNTIME (Q2)
\begin{table}
    \centering
    % \begin{tabular}{ll}
% \toprule
% Method & Runtime [s] \\
% \midrule
% TSMS & $1.87 \pm 1.87$ \\
% TSMS-St & $0.73 \pm 0.64$ \\
% TSMS-Per ~ & $2.43 \pm 2.08$ \\
% \bottomrule
% \end{tabular}

\begin{tabular}{l|lll}
\toprule
Method & ~ TSMS & ~ TSMS-St & ~ TSMS-Per \\
Runtime [s] ~~ & ~ $1.39 \pm 1.28$ & ~ $0.41 \pm 0.31$ & ~ $1.71 \pm 1.28$ \\
\bottomrule
\end{tabular}

\caption{Mean runtime (in seconds) plus/minus one standard deviation, measured over all datasets.}
\label{tab:runtimes}
\end{table}

In comparison to two different versions of \mname, namely \mname-St, which only creates the RoCs once at the start of inference and keeps static, as well as \mname-Per, where the RoCs are blindly recreated, we find that an informed, i.e., based on the detection of concept drifts, and infrequent recreation of the RoCs leads to better results in terms of predictive performance.
We observe that the periodic approach is only slightly better in terms of average ranking than the static approach, indicating that a blind recreation of RoCs is in most cases not worth the additional computational overhead.
%This can be explained by the excess of unnecessary RoC members being added due to noisy observations in the data, which does not make sense for the enrichment of RoCs since the underlying data did not experience any significant changes.
We additionally compare the runtime of each variant over a set of all datasets and report the average runtime, and its standard deviation, in \cref{tab:runtimes}.
Unsurprisingly, \mname-St is, on average, the fastest method since the RoC recreation procedure takes up most of the runtime.
The adaptive variant \mname\ falls in between the static and periodic variants, suggesting that the small trade in runtimes is worthwhile for improving predictive performance.
With these insights, we provide answers for research question \textbf{Q2}.

% EXPLAINABILITY ASPECTS (Q3)

\subsection{Explainability}
%To investigate the explainability of our method, we investigate \mname\ under three different aspects.
In this part, we answer research question \textbf{Q3} by showing how \mname\ supports explainability and covers multiple aspects of the forecasting task:

\begin{itemize}
    \item \textbf{A1}: Which input time series parts, i.e., lagged values, are more relevant for the prediction, and how this relevance evolves over time?
    \item \textbf{A2}: Why a given model $i$ is chosen to forecast the series at time $t$?
    \item \textbf{A3}:  Why a specific predicted value $\hat{x}^{(t+1)}$ is output by the selected model at time instant $t$?
    \item \textbf{A4}: What is the impact of concept drift occurrence on the pre-computed RoCs?
\end{itemize}

\begin{figure}
    \centering
    \includegraphics{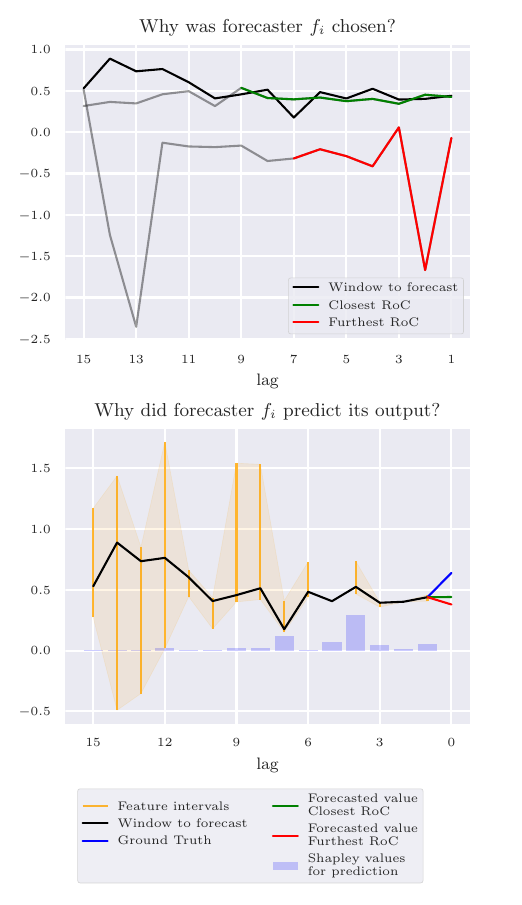}
    \caption{Visualization of \textbf{A2} and \textbf{A3} by investigating the closest and furthest RoC members, in addition to Shapley values and feature decision boundaries.}
    \label{fig:xai_why}
\end{figure}

\begin{figure}
    \centering
    \includegraphics{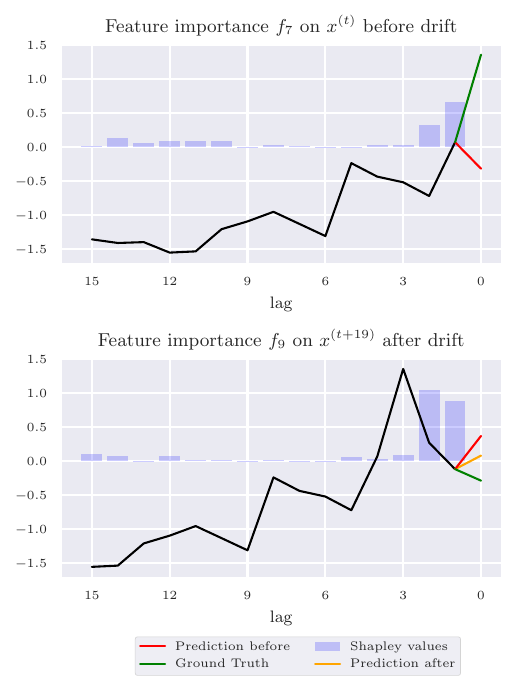}
    \caption{Comparison of Shapley values for prediction before and after concept drift where the chosen forecaster changes.}
    \label{fig:xai_change_importance}
\end{figure}

First, on the lower half of Figure \ref{fig:xai_why}, we can see the Shapley values of the $15$-time-lagged values used to forecast the true value of the time series marked in blue.
As mentioned earlier, by swapping the value function used for estimating Shapley values, we can explain both the achieved loss and the prediction itself.
For the prediction, the higher the bars (in blue), the more important the lag information fed to the forecasting model is.

As can be seen, many lags among the 15 lags seem to be irrelevant to predict the next value of the series. Surprisingly, this information is sparse and not strongly condensed around the most recent lags, in opposite to the assumption made by most of the traditional state-of-the-art models, including the Box-Jenkins ARIMA family of models \cite{box/etal/2005a}. This can be explained by the fact that these models assume the stationarity of the time series-generating process and the absence of concept drifts. So, once analyzed, lag information and contribution  in these models are kept fixed and restricted to the most three to four recent lagged observations \cite{Saadallah/Moreira/2018a}. However, this is not the case for many real-world scenarios. 
As can be seen in Figure \ref{fig:xai_change_importance}, the second most recent lag (lag 2) that used to have very low importance for the prediction before the occurrence of concept drift has become the most relevant one after the detection of drift. 
%\textbf{TODO: Is this still true? We show different models before and after drift dtection in figure 2: yep solved}
This shift in the importance of the input highlights the necessity of either triggering an update of the model using retraining or switching to another model that better handles the new dependence structure in the data after the occurrence of concept drift. Figure \ref{fig:xai_change_importance} shows that our framework ensures this update by switching to another model after drift detection, which predicts the true value of the series better. The prediction marked in red would be the forecast value if we stuck to the old model after drift detection, while the prediction marked in orange is the value output by the model selected by our method, which is closer to the true target value marked in green. Another potential insight provided by the tree-based models in relation to understanding the contribution of the lag information in the input time series to the prediction is provided by the paths taken in each tree. These paths enable the construction of boundaries around each lag via each node split criterion.
This means that changing the input value anywhere between the yellow bars in \cref{fig:xai_why} will have no effect on the models' prediction, which helps in understanding the robustness of the tree-based models. This knowledge is also of great help to forecasting practitioners as it helps them in conducting input change sensitivity analysis which is required in many forecasting financial and trading applications.
This addresses the explainability aspect raised in Question \textbf{A1}.

As previously explained, the reason for selecting forecaster $i$ at time $t$ is because its RoC contains a subseries that is the most similar to the current input pattern $\T{-L}{}\X{t}$ in terms of Dynamic Time Warping.
In Figure \ref{fig:xai_why} (upper half), we show the closest RoC marked in green to the current input pattern $\T{-L}{}\X{t}$. Its corresponding forecaster is thus chosen to forecast this data point. Examining the predictions of the model with the furthest RoC shown in red in Figure \ref{fig:xai_why} (lower half) and the prediction of our selected model marked in green validates our RoC-based selection as showing competence on patterns similar to the input pattern in question reflects the readiness of the model to process this pattern and output a prediction based on it. Hence, the large deviation between the RoC of the furthest model and the current pattern in Figure \ref{fig:xai_why} on the upper half is reflected in its prediction shown in the lower half. This answers the explainability aspect raised in Question \textbf{A2}. 
%In the same figure, on the right side, we show thre predicted value (in green) versus the true value (in red).
%Additionally, we visualize Shapley values for the predict.
%As mentioned earlier, by swapping the value function used for estimating Shapley values, we can explain both the achieved loss and the prediction itself.
%For prediction, the higher the bars (in blue) the more important this timestep was for the prediction.
%As can be seen, most time steps are (almost) irrelevant for the prediction.
%Additionally, since $\mathcal{P}$ only contains tree-based models, we can follow the paths taken in each tree and construct boundaries around each timestep via each nodes split criterion.
%That means that changing the input anywhere between the yellow bars will have no effect on the models prediction, which aids in understanding the robustness of the tree-based model.

\begin{figure}
    \centering
    \includegraphics{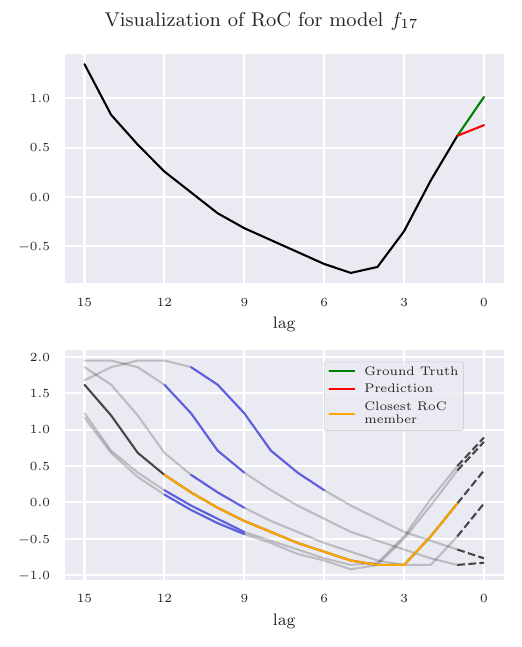}
    \caption{Visualization for $\mathcal{R}_{17}$}
    \label{fig:xai_roc_viz}
\end{figure}

Next, we provide some insights into how our method can help us to anticipate the model's output. This is one of the main important aspects of explainability as stated by Kim et al. \cite{kim2016examples}, who define explainability as the degree to which a human can consistently predict the model’s result. 
% In Figure \ref{fig:xai_roc_viz}, we retrieve the RoCs of the selected model that are similar to its current closest RoC to the current input pattern and their subsequent values, which are marked in a dotted line. 
In Figure \ref{fig:xai_roc_viz}, we visualize the RoC of the selected model $f_{17}$, including their subsequent values seen during RoC creation, which are marked in a dotted line.
A clear similarity between the output forecast by the model and the values that are subsequent to the RoCs sequences can be observed. The range of the subsequent values helps us in estimating the expected forecast value by the selected model on average. Comparing the predicted value by the model with this estimated average can be considered a debugging tool that helps us in detecting abnormal behavior of the model or check for significant changes in the data. This addresses the explainability aspect raised in Question \textbf{A3}.

\begin{figure}
    \centering
    \includegraphics{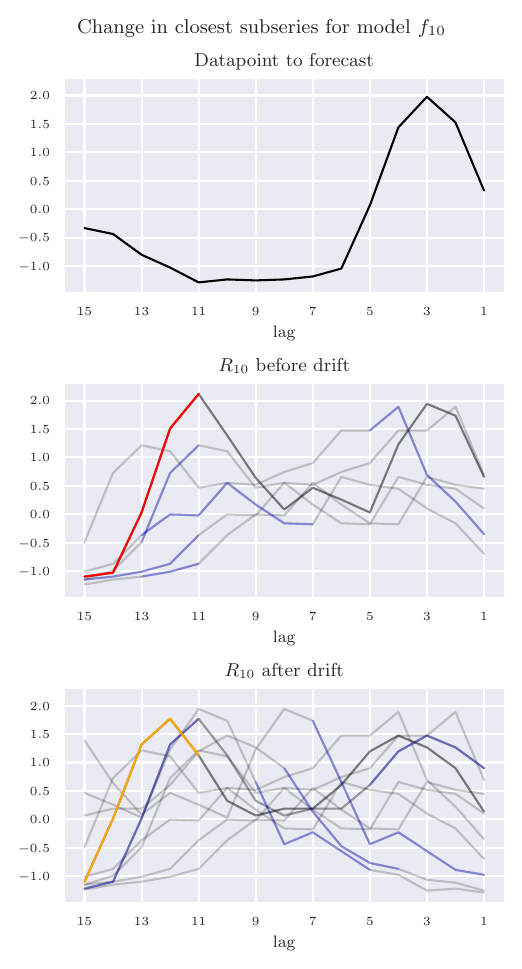}
    \caption{Comparison of $\mathcal{R}_{10}$ before and after a drift. Closest RoC member shown in orange and red, respectively.}
    \label{fig:xai_drift_model_no_change}
\end{figure}

To get a better understanding of the impact of drift detection, we investigate two scenarios.
First, shown in \cref{fig:xai_drift_model_no_change}, is a scenario where the drift detection was triggered but where the same model that would have been chosen without a drift was chosen regardless.
Nevertheless, as can be seen in a comparison between the middle and lower figures, in this case, the closest RoC member did, in fact, change.
The new closest RoC member (shown in orange) much more resembles the input in comparison to the previously closest RoC member (shown in red), indicating that the enrichment of the RoCs after the detected drift did adapt the RoCs to the time-changing data.

\begin{figure}
    \centering
    \includegraphics{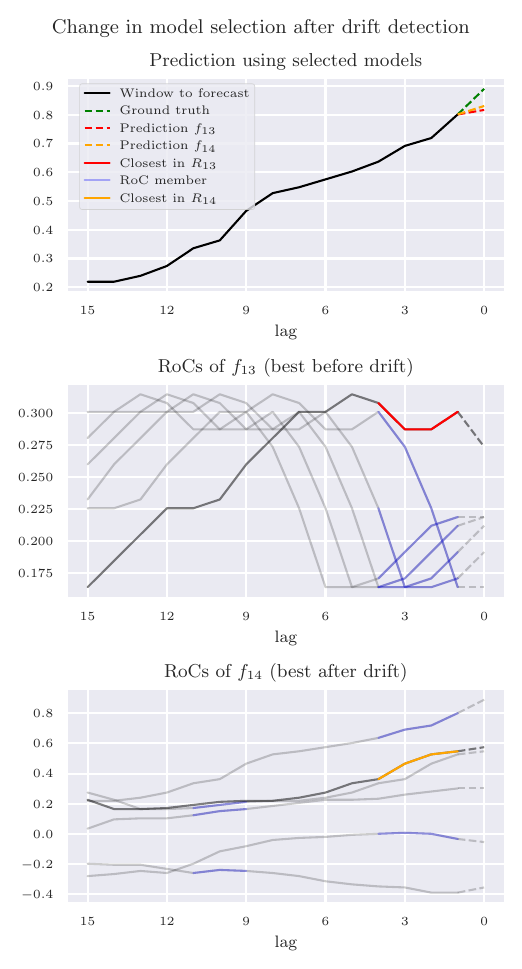}
    \caption{Comparison of $\mathcal{R}_{13}$ and $\mathcal{R}_{14}$ before and after a drift. Closest RoC member shown in orange and red, respectively.}
    \label{fig:xai_drift_model_change}
\end{figure}

Lastly, we consider a scenario where the chosen model did, in fact, change after a drift was triggered.
As can be seen in \cref{fig:xai_drift_model_change}, without enrichment of the RoCs, $f_{13}$ would have been chosen over $f_{14}$ in this case.
However, when investigating the prediction (upper figure), the prediction of $f_{16}$ (yellow) is closer to the ground truth value (green) compared to the prediction of $f_{13}$ (red).
Additionally, by investigating each model's RoC (middle and lower figure), it becomes clear that $f_{16}$ contains a member that resembles the input a lot more in comparison to the closest member in $\mathcal{R}_{13}$. These observations highlight the usefulness of the drift-aware adaptation of the RoCs over time. This answers the research question \textbf{A4}.

%By investigating the models' RoCs (before and after drifts), as well as investigating the individual predictions with Shapley values and looking at the robustness intervals for each feature, \mname\ provides a lot of insight into its decision making process, which clearly answer research question \textbf{Q3}.

\subsection{Discussion and future work}
The empirical results indicate that \mname\ performs very well in comparison to the tested baselines while simultaneously allowing insight into the decision-making process.
We argue that this combination is crucial when these methods are to be applied in real-world settings, especially in safety-critical applications where explainability and ML transparency are required.
We showcased, moreover, that \mname\ is able to adapt well to changes in the underlying time series data without the need to retrain every model in the pool.

In future work, we plan to extend our method to hybrid model pools by using the most efficient Shapley value estimation methods for each model family, such as TreeSHAP for tree-based models, DeepSHAP \cite{lundbergUnifiedApproachInterpreting2017} for Neural Networks, as well as KernelSHAP \cite{lundbergUnifiedApproachInterpreting2017} for remaining models.
Additionally, we plan to investigate the effect of other drift detection methods, especially in terms other than monitoring the deviation of the time series mean value.
Lastly, we want to enhance the explainability aspects further by reducing the RoCs number to the most important ones, which is highly important, especially for long time series where the number of RoCs is expected to grow very largely.

\section{Concluding Remarks}
This paper introduces \mname\, a novel method for online adaptive model selection on a pool of tree-based models.
Through the use of Shapley values, we are able to gain insight into its decision-making process, both for model selection, as well as for the individual model predictions.
We showed the advantages of \mname\ on \nds\ real-world datasets, both in terms of predictive performance as well as its explainability aspects. 

\section*{Acknowledgement}
This research has been funded by the Federal Ministry of Education and Research of Germany and the state of North Rhine-Westphalia as part of the Lamarr Institute for Machine Learning and Artificial Intelligence

\bibliographystyle{IEEEtran}
\bibliography{bibliography.bib}

\end{document}